\documentclass[sigconf]{acmart}
\renewcommand\footnotetextcopyrightpermission[1]{}
\AtBeginDocument{%
  \providecommand\BibTeX{{%
    \normalfont B\kern-0.5em{\scshape i\kern-0.25em b}\kern-0.8em\TeX}}}

\setcopyright{acmcopyright}
\copyrightyear{2022}
\acmYear{2022}
\acmDOI{XXXXXXX.XXXXXXX}

\acmConference[MM '22]{MM '22: }{10 – 14 Oct 2022}{Lisbon, Portugal}
\acmBooktitle{Multimedia 2022 Online, hosted by Lisbon, Portugal}
\acmPrice{15.00}
\acmISBN{978-1-4503-XXXX-X/18/06}

\usepackage{blindtext}
\usepackage{algorithmic}
\usepackage{algorithm,float}
\usepackage{graphicx}
\usepackage{comment}
\usepackage{multirow}
\usepackage{color,soul}
\usepackage{fancyhdr}
\usepackage{libertine}
\usepackage{amsthm}
\theoremstyle{definition}
\newtheorem{definition}{Definition}[section]

\usepackage{times}

\usepackage{soul}
\usepackage{url}
\usepackage[utf8]{inputenc}
\usepackage{graphicx}
\usepackage{amsmath}
\usepackage{booktabs}
\usepackage{multirow}
\usepackage{makecell}
\usepackage{adjustbox}
\usepackage{bm}

\begin{document}


%\title{Efficient Vertical Machine Unlearning: Selectively Removing Sensitive Information via Latent Feature Space}
\title{Efficient Attribute Unlearning: Towards Selective Removal of Input Attributes from Feature Representations} %junxiao updated 03/30


% \author{Anonymous authors}
% \author{Paper under double-blind review}
\author{
Tao Guo, Song Guo, Jiewei Zhang, Wenchao Xu, Junxiao Wang\\
Department of Computing, The Hong Kong Polytechnic University, Hong Kong, China
}


\renewcommand{\shortauthors}{Anonymous authors and Paper under double-blind review}

\begin{abstract}
%Retrieval of input attributes from the trained model has recently been shown to be feasible and pose a serious threat to user privacy.
%
%As a related, but easier attack compared to the full reconstruction of input data, the retrieval of input attributes can recover from deeper layers of a neural network the information even about attributes unrelated to the task at-hand, e.g. does a person that is recognized in a face recognition system wear a glass or the age range of that person.
%
Recently, the enactment of privacy regulations has promoted the rise of the machine unlearning paradigm. %concept.
%
%Existing studies of machine unlearning mainly focus on sample-wise unlearning, where data holders are entitled to proactively remove their data from a trained model at the sample level.
Existing studies of machine unlearning mainly focus on sample-wise unlearning, such that a learnt model will not expose user's privacy at the sample level.
%Existing studies of machine unlearning mainly focus on selectively removing input samples from feature representations such that a learnt model will not expose users' privacy at the sample-level.
%
Yet we argue that such ability of selective removal should also be presented at the attribute level, especially for the attributes irrelevant to the main task, e.g., whether a person recognized in a face recognition system wears glasses or the age range of that person.
%Yet we argue that privacy concern also arises at the attribute level, especially for the attributes irrelevant to the main task, e.g. whether a person recognized in a face recognition system wears glasses or the age range of that person.
%Yet we argue that such ability of selective removal should also be presented at the attribute-level, especially for the attributes that are unrelated to the main task, e.g. whether a person recognized in a face recognition system wears glasses or the age range of that person.
%
Through a comprehensive literature review, it is found that existing studies on attribute-related problems like fairness and de-biasing learning cannot address the above concerns properly. 
To bridge this gap, we propose a paradigm of selectively removing input attributes from feature representations which we name `attribute unlearning'.
In this paradigm, certain attributes will be accurately captured and detached from the learned feature representations at the stage of training, according to their mutual information.
%
% The particular attributes will be eventually eliminated if the training procedure is optimized towards convergence, while the rest of attributes related to the main task are preserved and contribute to a competitive model performance. 
The particular attributes will be progressively eliminated along with the training procedure towards convergence, while the rest of attributes related to the main task are preserved for achieving competitive model performance. 
Considering the computational complexity during the training process, we not only give a theoretically approximate training method, but also propose an acceleration scheme to speed up the training process.
We validate our method by spanning several datasets and models and demonstrate that our design can preserve model fidelity and reach prevailing unlearning efficacy with high efficiency.
The proposed unlearning paradigm builds a foundation for future machine unlearning system and will become an essential component of the latest privacy-related legislation.
\end{abstract} %junxiao updated 03/31


 %junxiao updated 03/30


\maketitle

\section{Introduction}

\begin{figure*}[t]
  \centering
  \includegraphics[width=0.85\linewidth]{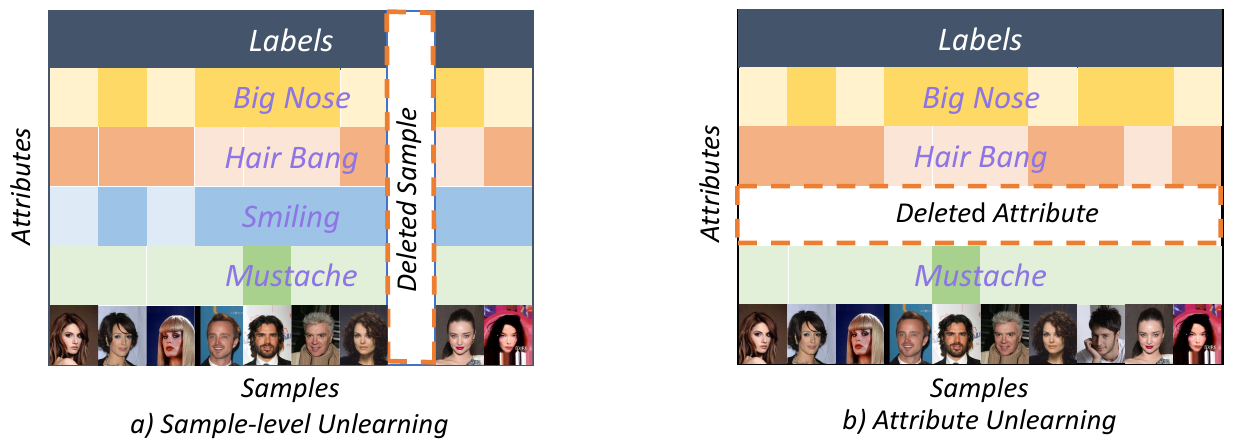}
  \caption{Inherent difference between proposed attribute unlearning and existing machine unlearning paradigm. The difference mainly lies in the objects of selective removal. Existing machine unlearning is aiming at selectively removing any influence of certain samples from learned feature representations. When existing machine unlearning removes input samples which contain certain attributes, the drawback is obvious: In addition to those attributes that are requested to be removed, some key attributes that contribute to the model performance have also been removed from learned feature representations.}
  \label{fig:MU}
\end{figure*}

% unlearning is necessary
% current unlearning is to delete data, i.e., the requirement from a certain user
% difficulties, limitations
% new definition of unlearning, in another dimension 
% benefits, not for a single user, but for a feature
% we propose something that is good

The progressive utilization of personal data has contributed to a wide spectrum of applications, e.g., intelligent healthcare, smart finance and secure access, etc. To prevent potential leakage and illegal use of private data, collaborative machine learning paradigms are proposed to train models without sharing users' data, e.g., federated learning \cite{yang2019federated}, split learning \cite{vepakomma2018split}. However, the models trained from isolated data may still have memories of them, leading to growing concerns of exposing sensitive information from existing trained models, such as diagnosis records, bank statement, facial images, etc. 

Recently, laws and regulations are enforced for companies and organizations on the `reasonable use of personal data', e.g., the European Union's General Data Protection Regulation (GDPR) \cite{mantelero2013eu} and the California Consumer Privacy Act (CCPA) \cite{web:lang:CCPA}, which allows people to get more control over their own data, and deliberately emphasize on selective removal of any personal information from learned feature representations \cite{bourtoule2021machine}. 
To proactively defend the privilege of data holders, machine unlearning paradigm has been brought up and sparked a boom both in academia and industry \cite{huang2021unlearnable,sekhari2021remember,gupta2021adaptive,brophy2021machine}. 

The most straightforward way of machine unlearning is to remove the data samples that are requested to be forgotten (referred to as target samples) from the original datasets, and retrain the ML model from scratch. Such way is easy to configure however can lead to prohibitive re-training costs, especially when the datasets are of large scale. 
%
% Retraining from scratch is easy to implement. However, when the amount of the original input samples is large and the model is complex, the computational costs of retraining is too large. 
To reduce the overhead, several alternative solutions have been proposed \cite{bourtoule2021machine,schelter2021hedgecut,neel2021descent,izzo2021approximate,wu2020deltagrad,graves2021amnesiac}.
%
%Wenchao: what is this sentence for?
% SISA \cite{bourtoule2021machine} works in an ensemble style, which is an efficient and general method to implement machine unlearning.
Nevertheless, status quo machine unlearning studies mainly focus on selectively removing input samples from feature representations such that a learnt model is able to remove user's private information only at the sample-level.
We argue that such ability of selective removal should also be presented at the attribute-level, especially for the attributes irrelevant to the main task, e.g., whether a person recognized in a face recognition system wears glasses or the age range of that person.
Although existing machine unlearning may remove all input samples which contain the certain attributes, the drawback is obvious: in addition to those attributes that are requested to be removed, some key attributes that contribute to the model performance may also be removed from learned feature representations (shown in figure~\ref{fig:MU}).

Existing studies on attribute problems, e.g. fairness and de-biasing learning \cite{singh2022anatomizing,ramaswamy2021fair,nam2020learning,wang2020towards}, are shown unable to address the above-mentioned issue. That's because,  
% Through a comprehensive literature review, it is found that existing studies on attribute-related problems like fairness and de-biasing learning \cite{singh2022anatomizing,ramaswamy2021fair,nam2020learning,wang2020towards} cannot well address the above concerns as well. %
the inherent objective of those studies is to learn fair and unbiased feature representations for the main task.
Therefore, those studies can help the trained model achieve better performance on the main task, but cannot ensure that particular attributes are eventually eliminated from the learned feature representations. 
In other words, status quo studies cannot offer the desired ability of selective removal at the attribute-level.
%
% How to efficiently removing input attributes from feature representations remains largely under-explored.

\noindent\textbf{Our Contributions.} In this paper, we bridge this gap by proposing a paradigm of selectively removing input attributes from feature representations which we name \textit{attribute unlearning}.
In this paradigm, certain attributes will be accurately captured and detached from the learned feature representations at the stage of training, according to their mutual information.
The particular attributes will be progressively eliminated along with the training procedure towards convergence, while the rest of attributes related to the main task are preserved for achieving competitive model performance. 
We focus on machine learning classification of images, and assume the attributes of the images have been well labeled in advance. 

We first define the problem of efficient attribute unlearning and show the challenges, formalization and goals. 
% compared with existing machine unlearning paradigm.
%
Specifically, we identify two unique challenges in the context of attribute unlearning from  the perspectives of embeddedness and interactivity of the attributes.
Identifying those challenges helps us better understand the problem and  clarifies the motivation of the paper.
We formalize the problem of attribute unlearning as a collaboration between two entities: 1) a trustful service provider with control over training process, data collection and management of whole model, 2) data holders, who provide individual data to service provider for training and validating.
Based on that formal definition and for purpose to better comply with efficient attribute unlearning, we dedicate the goals of the problem with four principles: 1) efficacy guarantee, 2) fidelity guarantee, 3) unlearning speedup, and 4) model agnostic.

Then, to achieve the above goals, we design an architecture of attribute unlearning with a proposed representation detachment approach.
We split the neural network into two parts, the bottom one is a representation detachment extractor, the top one is the classifier (see figure \ref{fig:arch}).
% not sure the sentences below is right
We leverage the intermediate representations extracted by the representation detachment extractor and use their mutual information to measure the correlation between the attributes from the latent feature space. 
Mutual information serves as an evaluation criteria here to quantify the mutual dependence and the expected `amount of information’.
To accurately capture and detach certain attributes from feature representations without demolishing the remaining key attributes, we dedicate a representation detachment loss, with that the information of certain attributes will be progressively removed from the learnt model at the stage of training, while the rest of attributes related to the main task are preserved for achieving competitive model.

% not sure the sentences below is right
In addition, considering that the aforementioned proposed loss function is difficult to optimize directly, and given that certain information is intractable in practice, we propose an upper bound approximation method to estimate the loss. To ensure the accuracy of the approximation, we show that the gap between the two is negligible via rigorous theoretical analysis.
To apply the approximate loss in practice, we use small auxiliary network to estimate required mutual information.
Moreover, considering the computational costs during the training process, we apply the local parallelism into our training architecture to reduce the severe time delay caused by the sequential order of backward propagation behaviors.
Given the intrinsic design of our architecture, auxiliary network for loss estimation can be used as module-wise propagation instead of global backward propagation.
Such that, the representation detachment extractor as well as the classifier can work together in parallel with a lower memory footprint and efficient speedup.

Finally, we conduct evaluation experiments based on an attribute unlearning task that trains on a deep neural network, where one or more certain attributes need to be removed or not use as required by custom.
We experiment our method with three commonly used datasets,
Fairface, CelebA, Cifar10, and nowadays most prevailing representative architecture, ResNet models.
The experimental results show that our method well performs in both aspect of unlearning efficacy, fidelity, and efficiency, as compared to two baseline methods.

To the best of our knowledge, we are the first to study machine unlearning at the attribute-level. We also envision the proposed unlearning paradigm builds a foundation for future machine unlearning system and will become an essential component of the latest privacy-related legislation.
% delete if needed
%\textit{Code and Data Sharing.}
% We will make the data snapshots and code used for attribute unlearning available to the research community in the hope that this will stimulate and facilitate further research.
%
To summarize, the main contributions of this paper are four-fold:
\begin{itemize}
\item We take the first step to investigate the machine unlearning at the attribute-level through defining the problem of efficient attribute unlearning including its challenges, formalization and goals.

\item We design an architecture of attribute unlearning with a proposed representation detachment approach. The particular attributes will be progressively eliminated along with the training procedure towards convergence, while the rest of attributes related to the main task are preserved for achieving competitive model performance.

\item Considering the computational complexity during the training process, we not only give a theoretically approximate training method, but also propose an acceleration scheme to speed up the training process.

\item We validate our method by spanning several datasets and models and demonstrate that our design can preserve model fidelity and reach prevailing unlearning efficacy with high efficiency.
\end{itemize}

\noindent\textbf{Roadmap.} In section~\ref{sec:Preliminaries}, we introduce the preliminaries of machine unlearning, and present the definition of our proposed attribute unlearning paradigm. We propose our representation detachment architecture including its approximate training and speedup in section~\ref{sec:Unlearning}. We conduct extensive experiments to illustrate the effectiveness of the proposed method in section~\ref{sec:Evaluation}. We discuss the related work in section~\ref{sec:Related} and conclude the paper in section~\ref{sec:Conclusion}.

\section{Attribute Unlearning Definition}\label{sec:Preliminaries}
For sake of completeness, we start the definition of attribute unlearning by briefly reviewing current emerging machine unlearning paradigm. Then we identify two challenges which is unique in the context of attribute unlearning, followed by presenting the formalization of attribute unlearning problem by revisiting challenges during the process. Towards overcoming these challenges, we also clarify the objectives that need to obtain in the end.

\subsection{Preliminaries of Machine Unlearning}
\noindent\textbf{Data Privacy Legislation.} The rapid development of big data boosts the advancement of deep learning by utilizing a massive amount of data generated from different sources across a wide range of sectors, such as healthcare and finance. However, the risk of privacy violation increases while we are savoring the benefits of `free' data.
Recently, numbers of privacy-preserving mechanisms \cite{papernot2016towards,liu2021machine} emerge for privacy protection at different stages from data generation to data processing.
Furthermore, privacy regulation \cite{mantelero2013eu,web:lang:CCPA,garcia2020lei,de2014protection} has been introduced to legitimize and protect sensitive information against hackers and leaks to make our digital environment safer and more secure. However, more and more emerging legislation emphasizes customers' rights other than merely keeping data safe from unauthorized access. Customers have rights to control the usage permission and know the purpose of using their information. 

\noindent\textbf{Studies on Machine Unlearning.}
To cope with the requirements in recently brought up legislation, e.g. the European Union's General Data Protection Regulation (GDPR) \cite{mantelero2013eu}, a novel branch of privacy-preserving machine learning paradigm arises, dubbed machine unlearning \cite{cao2015towards,bourtoule2021machine}.
Machine Unlearning is brought to address the challenges of unlearning required input samples from a learnt model. 
As the development of this field goes on, several research variants emerged. Some consider the enormous computational overhead during the process and focus on promoting unlearning efficiency \cite{huang2021unlearnable,sekhari2021remember,gupta2021adaptive,brophy2021machine}, another direction aiming at pushing this technique into extensive scenarios, e.g., recommendation system, federated learning, lifelong learning \cite{chen2022recommendation,wang2022federated,shibata2021learning}. 
Although achieving much progress, none of these works innovating from source, i.e., the objects of unlearning.

\subsection{Challenges in Attribute Unlearning}

To better understand the problem, we identify two unique challenges in the context of attribute unlearning from the perspectives of embeddedness and interactivity of the attributes. 

\noindent\textbf{Attributes are Embedded.}
% not sure the sentence below is right
Unlike the existing sample-wise machine unlearning paradigm where the target samples required to be removed are separable with each other in the input space.
Unlearning at the attribute-level does not have such advantage. In the context of attribute unlearning, the targets that need to be removed, i.e., certain attributes, are tightly embedded in the input space, which makes it difficult to separate them, especially for the unstructured data types, e.g., images and videos. Thus, it is challenging to find a general solution to this problem like existing machine unlearning which can directly retrain the models by deleting the target samples. 
The key reason is that those attributes will be entangled with each other from the view of input space. 

\noindent\textbf{Attributes are Interactive.} In addition to the embeddedness of the attributes, the interactivity of the attributes also poses the challenges.
During the training process, different attributes in the latent space will interact with each other, while those attributes will contribute to the final model performance in a complex manner. 
It is difficult to open the black-box and to understand that complex interactivity mode.
Accordingly, it is hard to conduct selective removal of certain attributes from the latent space and retain the other attributes intact. 
%
%Take facial recognition systems as an example, assume that data holders are concerned about the race attribute being exploited for malicious use, and the race attribute is closely related to the model main task.
%
In other words, the key challenge here in attribute unlearning is how to remove the influence of certain attributes to the large extent with the minimized collateral damage to the other attributes and model performance.

\begin{figure*}[t]
  \centering
  \includegraphics[width=0.85\linewidth]{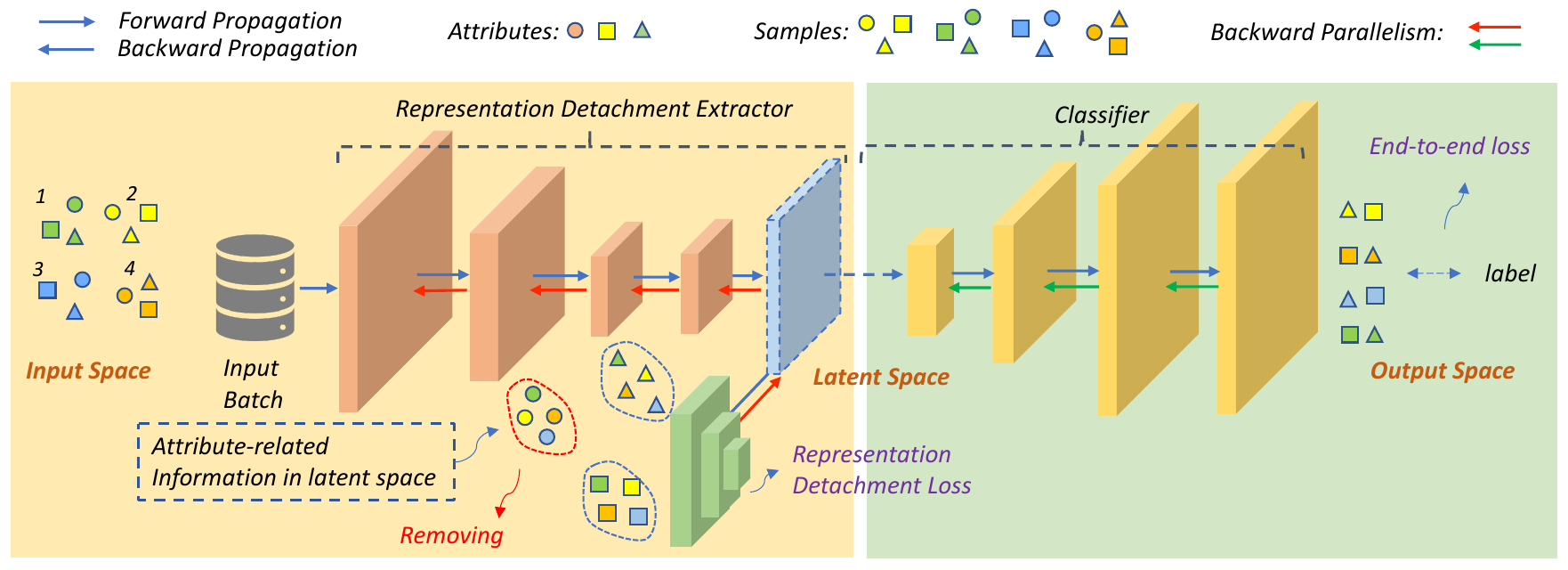}
  \caption{Architecture overview of proposed attribute unlearning. The neural network is split into two parts, the bottom one is a representation detachment extractor, the top one is the classifier. The intermediate representations are extracted by the extractor and used by a small auxiliary network to estimate the required mutual information and calculate a representation detachment loss function. That loss function will accurately capture the correlation between attributes from the latent feature space. 
  After iterative loss backward propagation and model update towards convergence, the information of certain attributes will be progressively removed from the learnt model, while the rest of attributes related to the main task are preserved for achieving competitive model performance.}
  \label{fig:arch}
\end{figure*}

\subsection{Formalization of Attribute Unlearning}
In this section, we formalize the problem of attribute unlearning as a collaboration between two entities: 1) a trustful service provider $\bm{S}$ with control over training process, data collection and management of whole model, 2) data holders, who provide individual data to service provider $\bm{S}$ for training and validating. 

Suppose there are $n$ data holders $\bm{C}=\{\bm{c}_1,\bm{c}_2,\dots,\bm{c}_n\}$, with $k$ attributes. Let $\bm{D}=\{\bm{d}_1,\bm{d}_2,\dots,\bm{d}_n\}$ be input space that model trains on, and $\bm{F}=\{\bm{f}_1,\bm{f}_2,\dots,\bm{f}_k\}$ be the latent feature space that input space can map to by $\bm{D} \rightarrow \bm{F}$, $\bm{\phi}$ be the model in hypothesis space after training. Within a finite time period, service provider $\bm{S}$ is required to return a model after unlearning of attributes $\{\bm{f}_u\}$.

\begin{definition}[\emph{Attribute Unlearning}]
 We define the algorithm $\bm{L}: \bm{F} \to \bm{\phi}$ be a learning process which maps latent feature space $\bm{F}$ into the hypothesis space of model $\bm{\phi}$. $\{\bm{f}_u\}$ be the attributes that data holders require to remove. We define a unlearning process $\bm{U}: \bm{L}(\bm{F}) \otimes \bm{F} \otimes \{\bm{f}_u\} \to \bm{\phi}$, whose input includes latent feature space $\bm{F}=\{\bm{f}_1,\bm{f}_2,\dots,\bm{f}_k\}$, a learnt model $\bm{L}(\bm{F})$ and the attributes $\{\bm{f}_u\}$ that required to be removed. Thus the objective of attribute unlearning problem can be described as
\begin{equation}
    \text{P}\Big[\bm{L}\big(\bm{F} \backslash \{\bm{f}_u\} \big)\Big] = \text{P}\Big[\bm{U}\Big(\bm{L}\big(\bm{F}\big), \bm{F}, \{\bm{f}_u\}\Big)\Big],
\end{equation}
where $\bm{F} \backslash \{\bm{f}_u\}$ defines the latent feature space without certain attributes $\{\bm{f}_u\}$. $\text{P}(\bm{\cdot})$ denotes the distribution of feature representations in the model output.

% Assuming that we are solving the problem of a supervised learning task, we also need to satisfy the task performance after unlearning selective features from the learnt model. 
\end{definition}
% junxiao updated 04/04

%We also depict the architecture of sample learning and attribute learning in Fig.~\ref{fig:MU} to better understand the difference between the two modes. We can see from the figure that the horizontal axis represents the input space, the vertical axis represents latent feature space and each sample contains several latent features inside the data. The comparison manifests the intrinsic difference between them is that the sample unlearning aiming to unlearn models by deleting certain data samples while attribute unlearning targets on the attributes. 

% \begin{figure}[t]
%   \centering
%   \includegraphics[width=1.0\linewidth]{figures/VMU1.pdf}
% ~  \caption{\textbf{Vertical Machine Learning}: Edge devices with limited resources leverage powerful computation resources on cloud, but may introduce information leakage when encountering malicious attacker. }
%   \label{fig:VMU}
% \end{figure}

\subsection{Principles in Attribute Unlearning}\label{subsec:Principles}
To better illustrate the goals of proposed problem, we dedicate several principles that efficient attribute unlearning needs to meet:

\textbf{P1 Efficacy Guarantee.} First, the optimal strategy to implement the attribute unlearning paradigm should eliminate as much as possible the model's memory of the target attributes. 
To measure the a strategy's performance under this principle, we will also introduce a general metrics for that in section~\ref{subsec:metrics}.

\textbf{P2 Fidelity Guarantee.} Secondly, since attributes interact with each other during training, there is a trade-off between how well the model unlearns and the fidelity it retains. 
The best strategy to implement the attribute learning paradigm should maintain a competitive model performance as much as possible while eliminating the model's memory of the target attributes. 

\textbf{P3 Unlearning Speedup.}
\label{goal:G3}
Thirdly, attribute learning should consider unlearning efficiency. In addition to getting decent performance, the optimal strategy should also have as little time to unlearn as possible. In line with this principle, we propose an acceleration strategy to speed up the unlearning process (as referred to section~\ref{subsec:Acceleration}). 

\textbf{P4 Model Agnostic.} 
Fourthly, the ideal strategy to implement the attribute unlearning should be general across different model architectures, which means that the aforementioned goals ought to be satisfied by any model when performing that unlearning strategy. 

% junxiao updated 04/04

\section{Efficient Attribute Unlearning}\label{sec:Unlearning}

%\section{Representation Disentanglement}
Kept those challenges and goals in mind, we design an architecture of attribute unlearning with a proposed representation detachment approach. 
In this section, we start from introducing the design and workflow of proposed representation detachment. 

\subsection{Representation Detachment Overview}
The overall architecture of our attribute unlearning process with representation detachment approach is described in figure~\ref{fig:arch}. 
%We first explain the training procedure, then we will describe the inference period after the unlearnt model being fixed.
%Specially, we consider an image classification problem during the training process. 

%In accord with aforementioned attribute unlearning problem, we view the problem with two parities, service provider $\bm{S}$ who owns a model $\bm{M}$, and data holders who provide input samples $\bm{D}=\{\bm{d}_1,\bm{d}_2,\dots,\bm{d}_n\}$ in the input space with corresponding $k$ attributes $\bm{F}=\{\bm{f}_1,\bm{f}_2,\dots,\bm{f}_k\}$ in the latent feature space. Given the intrinsic privilege of controlling and protection over individual data, data holders may request the model to remove selected attributes $\{\bm{f}_u\}$ and obtain model $\bm{M}_u$. %We address this issues by applying our representation detachment approach.

\noindent\textbf{Model Agnostic Design.}
In the proposed architecture, the neural network is split into two parts, the bottom one is a representation detachment extractor, the top one is the classifier.
We adopt the design idea of bypass, which is widely used in many popular neural network architectures like ResNet.
Without changing the original model architecture, we place a small auxiliary network between the feature extractor and classifier, so that the original model and the newly added auxiliary network can work together to complete the task of attribute unlearning.
It is noted that the proposed architecture meets the model agnostic design, which has been emphasized in section~\ref{subsec:Principles}. 

\noindent\textbf{Overall Workflow.}
The representation detachment approach is realized through the forward propagation and backward propagation with the collaboration between the original model and the auxiliary network.
%We
We focus on machine learning classification of images, and assume the attributes of the images have been well labeled.
First, the input data goes into the feature extractor, and through the forward propagation, the intermediate representations are extracted by the extractor and then are used by the auxiliary network to estimate the required mutual information and calculate a representation detachment loss function. 
That loss function will accurately capture the correlation between attributes from the latent feature space. 
During the backward propagation of that loss, the bottom-part network structure, i.e. the representation detachment extractor will be updated. 
The loss backward propagation of the original model will only update the top-part network structure, i.e. the classifier.
After iterative loss backward propagation and model update towards convergence, the information of certain attributes will be progressively removed from the learnt model, while the rest of attributes related to the main task are preserved for achieving competitive model performance.

\subsection{Auxiliary Network}\label{subsec:Auxiliary}
The auxiliary network is a small network consisting of few convolutional layers and linear layers.
As compared to the original model, the additional computation and storage overhead is limited. 
The input of the auxiliary network include (1) the extracted intermediate representations, a.k.a. feature map, and (2) the labeled attributes. 
The output of the auxiliary network is a loss score which is computed by a representation detachment loss function.
During the backward propagation of that loss, the bottom-part network structure will be updated.
Target attributes will also be selectively removed from the feature representations during that process.

\subsubsection{Representation Detachment Loss}
Given feature map $\bm{h}$ corresponding to the input data $\bm{x}$, we use the mutual information $\textbf{I}(\bm{h},\bm{x})$ and $\textbf{I}(\bm{h},\bm{y})$ to measure the amount of all retained information from $\bm{x}$ and task-relevant information in $\bm{h}$.
We use $\bm{y}$ to represent the labels related to the main task identification.
We denote by $\bm{s}$ as the selected attribute related information.
%We denote by $\bm{r}$ and $\bm{s}$ the nuisance attributes and the target attributes related information, respectively. 
%
We dedicate a representation detachment loss function for the auxiliary network, whose optimization objective is to eliminate the learned information in $\textbf{I}(\bm{h},\bm{s})$ while in the meantime preserving other information in $\textbf{I}(\bm{h},\bm{x})$.
The formal definition of that loss function is given below,
%
%\begin{equation} 
%\label{original loss func}
%\begin{split} 
    %\mathcal{L}&=\alpha\big[-\textbf{I}(\bm{h},\bm{x})+\beta \textbf{I}(\bm{h}, \bm{r}^*)+\gamma \textbf{I}(\bm{h}, \bm{s}^*)\big], \\
    %\bm{r}^*&=\argmax\bm{r},\textbf{I}(\bm{r},\bm{x})>0, \textbf{I}(\bm{r},\bm{y})=0\textbf{I}(\bm{h},\bm{r})
%\end{split}
%\end{equation}
%
\begin{equation}
     \begin{split}
         \mathcal{L}&=\alpha\bm{\cdot}\big[-\textbf{I}(\bm{h},\bm{x})+\gamma\bm{\cdot}\textbf{I}(\bm{h}, \hat{\bm{s}})\big], \\
         %\hat{\bm{r}}&=\arg\max_{\bm{r}}\textbf{I}(\bm{h},\bm{r}), \ s.t. \ \textbf{I}(\bm{r},\bm{x})>0,\textbf{I}(\bm{r},\bm{y})=0, \\
         %\hat{\bm{s}}&=\arg\max_{\bm{s}}\textbf{I}(\bm{h},\bm{s}), \ s.t. \ \textbf{I}(\bm{s},\bm{x})>0,\textbf{I}(\bm{s},\bm{y})=0,
         \mathrm{s.t.} \quad \hat{\bm{s}}&= {\arg_{\bm{s}}\max}_{\textbf{I}(\bm{s}, \bm{x})>0, \textbf{I}(\bm{s},\bm{y})=0} \textbf{I}(\bm{h},\bm{s}), 
     \end{split}
\label{equ:loss}
\end{equation}
where $\alpha$, $\gamma>0$ are hyperparameters. We will further investigate their impacts in section~\ref{sec:Evaluation}. 
$\textbf{I}(\bm{a},\bm{b})$ represents the mutual information between variable $\bm{a}$ and $\bm{b}$.
Specially, the term $\textbf{I}(\bm{h},\bm{x})$ is to capture the mutual information between feature map and input image.
%
%The term $\textbf{I}(\bm{h}, \hat{\bm{r}})$ is to extract the mutual information between feature map with task-irrelevant nuisance, like background, which serves as the purpose of improving model performance.
%
The term $\textbf{I}(\bm{h}, \hat{\bm{s}})$ aims to extract the mutual information between feature map with selected attributes, for the purpose of eliminating attribute influence. 
%
%Mutual information serves as an evaluation criteria here as a measure of mutual dependence between the two variable in the probability or information theory and first brought up by `Claude Shannon' in \cite{shannon2001mathematical}. And in machine learning field, mutual information is linked to the entropy of a random variable and quantify the expected `amount of information' in a given variable \cite{batina2011mutual,gierlichs2008mutual}.

%\noindent\textbf{Mutual Information Estimation.} 
%It is noted that the desirable mutual information $\textbf{I}(\bm{h},\hat{\bm{r}})$ and $\textbf{I}(\bm{h},\hat{\bm{s}})$ are hard to obtain in practice.
%
%As shown in proposition~\ref{approximate}, we use their upper bound instead to express the information we capture as much as possible in $\bm{h}$ during training.

\subsubsection{Loss Approximation}
It is noted that the desirable mutual information %$\textbf{I}(\bm{h},\hat{\bm{r}})$ and
$\textbf{I}(\bm{h},\hat{\bm{s}})$ is hard to obtain in practice.
According to lemma~\ref{lem:approximate}, we suggest to use the upper bound of that loss for an effective approximation.
We also show that the gap between the two is theoretically small with our lemma~\ref{lem:gap}.
\begin{lemma} \label{lem:approximate}
Suppose the Markov chain $(\bm{y},\bm{s})\rightarrow \bm{x}\rightarrow \bm{h}$ holds. Then the upper bound of $\mathcal{L}$ is as follows
\begin{eqnarray}
    \mathcal{L} \le -\lambda_1\bm{{\rm I}}(\bm{h},\bm{x})-\lambda_2\bm{{\rm I}}(\bm{h},\bm{y})-\lambda_3\bm{{\rm I}}(\bm{h},\bm{z}) 
    =\overline{\mathcal{L}},
\end{eqnarray}
where $\lambda_1=\alpha(1-\beta)$, $\lambda_2=\alpha\beta$, $\lambda_3=\alpha(\beta-\gamma)$, $\beta\geq\gamma$, and $\bm{z}$ represents the labels related to the target attribute identification.
\end{lemma}
\begin{lemma} \label{lem:gap}
Assume there exists a deterministic function map $\bm{x}$ to $\bm{y}$, the gap $\epsilon=\overline{\mathcal{L}}-\mathcal{L}$ is bounded by
\begin{equation}
    \epsilon \le \alpha\beta\bm{\cdot}\big[\bm{{\rm I}}(\bm{x},\bm{y})-\bm{{\rm I}}(\bm{h},\bm{y})-\bm{{\rm I}}(\bm{h},\bm{z})\big].
\end{equation}
\end{lemma}
As for the multi-attribute case, it can be extended as $\epsilon \le \alpha\beta(I(\boldsymbol{x},y)-I(\boldsymbol{h},y)-\sum I(\boldsymbol{h},z_i))$.
\noindent The detailed proof for lemma~\ref{lem:approximate} and \ref{lem:gap} is included in the appendix.

\subsubsection{Mutual Information Estimation} 
To apply the above loss in practice, we use the auxiliary network to estimate them. 
Assuming $\textbf{R}(\bm{x}|\bm{h})$ denotes the expected error for reconstructing $\bm{x}$ from $\bm{h}$. 
According to the widely known conclusion of mutual information estimation \cite{Wang2021revisiting,hjelm2018learning,makhzani2015adversarial}, we can estimate $\textbf{I}(\bm{h},\bm{x})$ by training a decoder parameterized by $\omega$ to obtain the minimal reconstruction loss. In our implementation we use binary cross-entropy loss for $\textbf{R}_\omega(\bm{x}|\bm{h})$.

As for $\textbf{I}(\bm{h},\bm{y})$ and $\textbf{I}(\bm{h},\bm{z})$, given that definition of mutual information, we have  $\textbf{I}(\bm{h},\bm{y})$=$ \textbf{H}(\bm{y})$-$\textbf{H}(\bm{y}|\bm{h})$=$ \textbf{H}(\bm{y})$-$\textbf{E}_{(\bm{h},\bm{y})}[-\log \textbf{P}(\bm{y}|\bm{h})]$, where $\textbf{H}(\bm{\cdot})$ represents the marginal entropy, and $\textbf{E}$ represents the  expectation.
Then, we can implement and train two auxiliary classifiers to approximate $\textbf{I}(\bm{h},\bm{y})$ and $\textbf{I}(\bm{h},\bm{z})$, respectively.
When training those classifiers, we use a contrastive loss by pulling the same class together and different class apart for better performance, which has been widely exploited in recent arisen contrastive learning approaches \cite{khosla2020supervised,chen2020simple,tian2020makes}.
We will also prove with comprehensive experiments that our proposed method meets the principles of efficacy guarantee and fidelity guarantee as referred in section~\ref{subsec:Principles}.

%Inspired by recent arisen contrastive learning \cite{khosla2020supervised,chen2020simple,tian2020makes}, we adopt supervised contrastive learning loss in \cite{khosla2020supervised} instead of traditional cross-entropy for training above two auxiliary classifiers, where label information is leveraged to pull the same class together and different class apart for better performance.

%given that $\textbf{I}(\bm{h},\bm{y}) = \textbf{H}(\bm{y})-\textbf{H}(\bm{y}|\bm{h}) = \textbf{H}(\bm{y})-\textbf{E}_{(\bm{h},\bm{y})}[-\log p(\bm{y}|\bm{h})]$, we train auxiliary classifiers $q_{\psi_1}(y|\boldsymbol{h})$, $q_{\psi_2}(z|\boldsymbol{h})$ to approximate.
\subsection{Parallelism Acceleration}\label{subsec:Acceleration}
To better accord with the unlearning speedup principle which has been emphasized in section~\ref{subsec:Principles}, we apply the local parallelism into our training architecture to reduce the severe time delay caused by the sequential order of backward propagation behaviors.

\noindent\textbf{Where to Split the Network.}
Before investigating the acceleration, we need to recall that our architecture leverages the intrinsic advantage by splitting the original model into two parts, one is a representation detachment extractor $\Phi$ on the bottom, the other is a classifier $\Omega$ on the top, as illustrated in figure~\ref{fig:arch}. 
According to some recent findings \cite{ren2021interpreting,li2021visualizing}, feature map of lower complexity orders represents more general information, while that of higher complexity orders give more noise. According to this inspiring discovery, we are motivated to use the feature map of the middling complexity as the input of our auxiliary network. In other words, we will split the original model from a middle range layer. 
This split separates the computational load during training into two parts, which creates the possibility for parallel acceleration.

\noindent\textbf{Backwards Locking Problem.}
Although the network is explicitly divided into two parts, their parallel acceleration still presents other challenges.
According to the studies in \cite{laskin2020parallel,huo2018decoupled,jaderberg2017decoupled}, abide by the rule of end-to-end method will result in backwards locking when conducting loss backward propagation and model update.
This means that no modules can be updated before their predecessor modules have executed both forward and backward propagation. 
Due to those sequential order of backward propagation behaviors, there is a serious delay during the training process.
We manage to address the problem by bringing local parallelism into our architecture.

\noindent\textbf{Local Parallelism.}
In our implementation, both the representation detachment extractor and the classifier conduct their own backward propagation and model update in their own part.
Such that, the auxiliary network for mutual information estimation is used as module-wise propagation instead of global backward propagation.
The representation detachment extractor and the classifier can work together in parallel with a lower memory footprint and efficient speedup.
In section~\ref{sec:Evaluation}, we will provide an efficiency analysis of our approach against end-to-end method. 

\section{Evaluation}\label{sec:Evaluation}
To evaluate the effectiveness of our method in the attribute unlearning scenario, we propose some general evaluation metrics and evaluate them across several datatsets and models. 

\subsection{Experiments Setting}
% The evaluation scenario is based on a unlearning task that trains on a deep neural network, where one or more sensitive features need to be deleted or not use as required by customs. However, the model performance for the main task should be preserved. 

\textbf{Datasets.} We experiment our method with three widely used datasets, Fairface \cite{karkkainen2019fairface}, CelebA \cite{liu2018large}, Cifar10 \cite{krizhevsky2010convolutional}, and for each datasets we select corresponding task and attribute to unlearn. Fairface contains 108,501 balanced images with race as the unlearn attribute and gender as the task attribute. Cifar10 consists of 202,599 celebrity faces with 40 attribute annotation. In our experiment we select Male as the unlearn attribute and smiling as task. Cifar10 contains 50,000 training data and 10,000 validation data, the attribute unlearn here is living/non-living. Experiments are implemented using Pytorch with Nvidia 3080 GPU. The backbone network is nowadays most prevailing representative architecture, ResNet models, e.g., ResNet18, ResNet34, ResNet50 \cite{he2016deep}.
% which satisfies our need by having sensitive latent feature that we can unlearn. For all the experiments, we use nowadays most prevailing representative architecture, ResNet models, ,e.g., ResNet-18, ResNet-34, ResNet-50, as the backbone model to unlearn from.

\begin{figure}[t]
  \centering
  \includegraphics[width=1.0\linewidth]{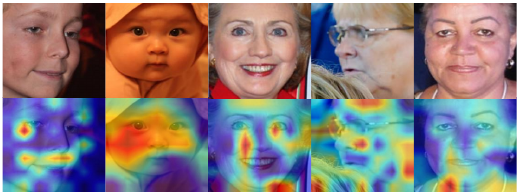}
  \caption{Attention results with Gradcam \cite{selvaraju2017grad}. The first row shows the original input and the second row shows raw input data and with Gradcam's attention on it by given certain attribute. With the attention being captured, we will remove the key area and generate a new dataset without the selected attributes for baseline retraining.}
  \label{fig:gradcam}
  \vspace{2mm}
\end{figure}

\noindent\textbf{Baseline.} We choose two baselines for comparison of our approach and use them as the upper bound and lower bound for unlearning efficacy and fidelity retained separately. Baseline 1 is the learnt model before unlearning, so in this case, the fidelity of baseline 1 should be optimal and the efficacy should be the worst, since none of the information has been removed from the model.
For baseline 2, apparently, deleting selected attribute directly from input samples and using the remaining for retraining is the most straightforward way for unlearning such attribute. However, it is difficult to execute since attributes are embedded and mixed together inside the input images. Thus we use GradCam \cite{selvaraju2017grad} to locate the attention area of the attribute and remove the key area, as shown in figure~\ref{fig:gradcam}.

\subsection{Evaluation Metrics}\label{subsec:metrics}
In unlearning problem setting, the goal is to cleanly remove the selected attributes and preserve the model performance as possible. Thus we set up a series of evaluation metrics to assess the contribution of our method to attribute unlearning from four key aspects.

\begin{figure}[t]
  \centering
  \includegraphics[width=1.0\linewidth]{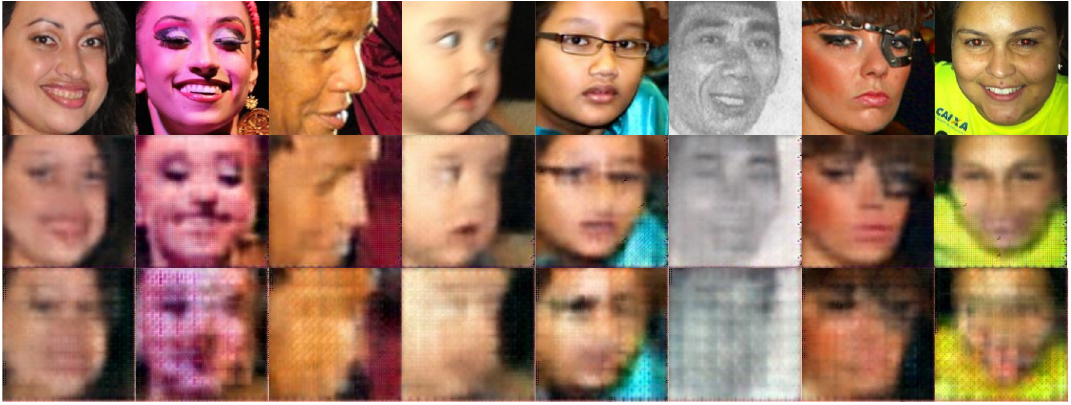}
  \caption{Attribute reconstruction results. The first row shows raw input data. The second row and the third row show the reconstructed input images from the learnt model before and after unlearning, respectively. We can see that before unlearning, attributes can be recovered to a large extent. While after unlearning, certain area is hard to recognized with bare-eyes.}
  \label{fig:recon}
  \vspace{2mm}
\end{figure}

\noindent\textbf{Unlearning Efficacy.} We consider an attack setting with parities of two, a model provider and a malicious attacker. Suppose $(\bm{h},\bm{z})$ pairs represent the feature map and attributes. The attacker leverages $(\bm{h},\bm{z})$ pairs to train a decoder with ResNet18 as its backbone, and in attempting to infer the unlearnt attribute from the feature map. 
So we focus on the prevention of sensitive attribute leakage and use the prediction accuracy of sensitive attribute as the indicator for the efficacy of attribute unlearning. The lower the value, the better the unlearning works. We also visualize the efficacy by recovering the input images, as shown in figure ~\ref{fig:recon}.
%and compare the reconstructed images with the original ones in viewing the difference of sensitive feature.

\noindent\textbf{Fidelity Retained.} In spite of obtaining high unlearning efficacy, we also need to retain the performance of the main task of the trained model as much as possible, which we called fidelity retained. We use top-1 validation accuracy on the main task as our measurement for this factor. Take Fairface dataset as an example, top-1 accuracy for prediction of gender signify the fidelity capability of the model. Higher accuracy means better utility retained after experiencing that unlearning process.

\noindent\textbf{Unlearning Efficiency.} The efficiency of unlearning serves as a novel indicator which aims at introducing as less additional overhead as possible when performing the unlearning process. Here we use unlearning latency as the metrics to evaluate efficiency.

\label{AUM}
\noindent\textbf{Attribute Unlearning Measure.} Other than the above metrics, we introduce a new evaluation metric \textit{Attribute Unlearning Measure (AUM)}. AUM is calculated as the harmonic mean of the aforementioned efficacy and fidelity, as following
\begin{equation}
    \text{AUM} = \frac{2\times \text{Efficiency Score} \times \text{Fidelit Score}}{\text{Efficiency Score} + \text{Fidelity Score}}.
\end{equation}
Here we use this metric to unify a single indicator to combine efficacy and fidelity, the higher the better.
%These measure is first used in \cite{shibata2021learning} for the forgetting problem in lifelong learning. Here we use this metric to unify a single indicator to combine efficacy and fidelity, the higher the better.

\begin{table*}[!h]
\renewcommand\arraystretch{1.1}
\centering
%\begin{adjustbox}{width=0.5\textwidth}
\begin{tabular}{l|l|l|l|l|l|l|l|l|l|l}
\Xhline{3\arrayrulewidth}
\multirow{3}*{Datasets}&\multirow{3}*{Models}&\multicolumn{3}{c|}{Baseline 2}  &\multicolumn{3}{c|}{Attribute Unlearning}  &\multicolumn{3}{c}{Baseline 1} \\
\cline{3-11}
~& &Efficacy&Fidelity&AUM&Efficacy&Fidelity&AUM& Efficacy&Fidelity&AUM  \\ 
~ & &(\textit{optimal}$\ast$)&(\textit{baseline} \textendash)& ($\uparrow$ \textit{better}) & ($\downarrow$ \textit{better}) & ($\uparrow$ \textit{better})& ($\uparrow$ \textit{better})&(\textit{baseline}\textendash)&(\textit{optimal}$\ast$)& ($\uparrow$ \textit{better}) \\ \Xhline{3\arrayrulewidth}

\multirow{3}*{Fairface}&ResNet18 &0.1421&0.5810&0.6928 &0.1352 & 0.7950& \textbf{0.8284} & 0.3107& 0.8330&0.7544  \\  
~&ResNet34&0.1325&0.5801&0.6953 &0.1350  & 0.7988& \textbf{0.8306} & 0.2844& 0.8340& 0.7703  \\ 
~&ResNet50&0.1367&0.5730&0.6889 &0.1365 &  0.8077&\textbf{0.8347} & 0.2977& 0.8470&0.7679\\ \hline
   
\multirow{3}*{CelebA}&ResNet18&0.4989& 0.5010&0.5144 & 0.5142 & 0.7830&\textbf{0.5994} & 0.6391& 0.7901&0.4955 \\  
~&ResNet34&0.5077& 0.5010& 0.4966& 0.5035 &  0.7930&\textbf{0.6107}  & 0.6010& 0.8032&0.5332 \\ 
~&ResNet50&0.4976& 0.5110& 0.5067 & 0.5082 &  0.8101&\textbf{0.6120}  &0.6572& 0.8121&0.4821 \\ \hline
   
\multirow{3}*{Cifar10}&ResNet18& 0.1030& 0.5443&0.6775 &0.1095 &0.9351&\textbf{0.9123} & 0.2668&0.9610&0.8318  \\  
~&ResNet34& 0.1010& 0.5070&0.6484 & 0.1057 & 0.9360&\textbf{0.9147} & 0.2357&0.9632&0.8523 \\  
~&ResNet50& 0.0993& 0.5130&0.6537 &0.0982  & 0.9452&\textbf{0.9230} & 0.3226&0.9681&0.7971 \\ \Xhline{3\arrayrulewidth}
    
\end{tabular}
%\end{adjustbox}
\vspace{1mm}
\caption{Unlearning efficacy vs. fidelity retained. Baseline 1 and baseline 2 represents `before unlearning' and `retraining after key attributes area removal' respectively. The third and tenth columns represent optimal efficacy and fidelity, one can observe that our methods approach both optimal results. Results of AUM outperform both baselines.}
\label{tab:utility}
\end{table*}

\begin{table*}[!h]
\centering
\begin{tabular}{l|l|l|l|l|l|l|l|l|l|l}
\Xhline{3\arrayrulewidth}
\multirow{3}*{Models}&\multicolumn{2}{c|}{Unlearning Latency}  &\multicolumn{2}{c|}{Efficacy}  &\multicolumn{2}{c|}{Fidelity}&\multicolumn{2}{c|}{Memory Consumption}&\multicolumn{2}{c}{Speedup Ratio}\\

~&\multicolumn{2}{c|}{ ($\downarrow$  \textit{better})} & \multicolumn{2}{c|}{($\downarrow$  \textit{better})}& \multicolumn{2}{c|}{($\uparrow$  \textit{better})}& \multicolumn{2}{c|}{($\downarrow$  \textit{better})} & \multicolumn{2}{c}{($\uparrow$  \textit{better})}
\\ \cline{2-11}
~& Raw&Speedup & Raw&Speedup & Raw&Speedup & Raw&Speedup &  Raw&Speedup \\

\Xhline{3\arrayrulewidth}
ResNet18 &  32.297 &  24.265 & 0.1039  & 0.0996 & 0.9301  & 0.9351 &281.01 & 231.49 \textit{(18.8$\%\downarrow$)}& \textendash &1.331$\times$   \\ \hline
ResNet34  & 45.580  & 34.308  &  0.1164 & 0.1057  & 0.9297  &  0.9360  &416.03 & 327.49 \textit{(21.3$\%\downarrow$)}& \textendash &1.328$\times$   \\ \hline
ResNet50  & 63.632  & 44.790 & 0.1129 & 0.0982  & 0.9436  & 0.9452 &546.17 &438.87 \textit{(19.6$\%\downarrow$)} &\textendash &1.421$\times$    \\
\Xhline{3\arrayrulewidth}
\end{tabular}
\vspace{1mm}
\caption{Unlearning efficiency and memory footprint results.  Attribute unlearning after acceleration express efficiency and low memory footprint. Latency speed up by 1.42$\times$ and memory consumption lower by 21.3\% with comparative efficacy and fidelity.}
\label{tab:efficiency}
\end{table*}

\subsection{Comparative Results}
\begin{figure}[t]
  \centering
  \includegraphics[width=1.0\linewidth]{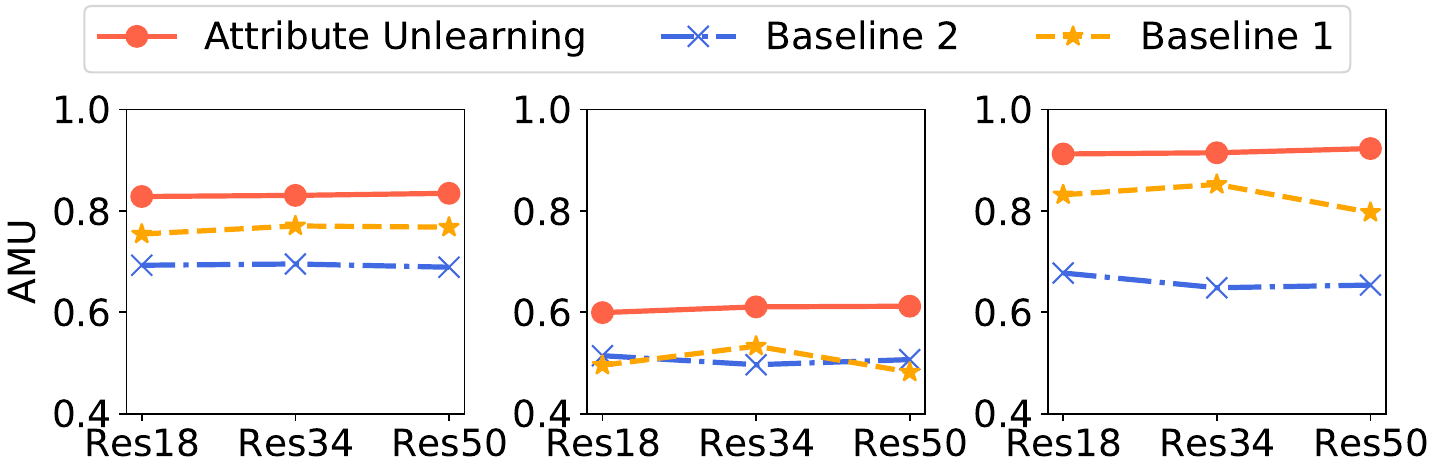}
  \caption{Attribute unlearning measure (AMU) results. Left, center and right subfigures are with Fairface, CelebA and Cifar10 datasets. AUM of our method outperforms the other baselines spanning datasets and models.}
  \label{fig:amu}
  \vspace{2mm}
\end{figure}

\noindent\textbf{Efficacy and Fidelity} Table~\ref{tab:utility} and figure~\ref{fig:amu} show comparative results of our method. We can see that our method is the best across all datasets and models in AUM against both baselines. It can also be observed that our method approaching optimal fidelity and optimal efficacy at the same time against to column 3 and column 10, which means that our method can eliminate the selected attributes and preserve the model performance simultaneously. Our method can perfectly satisfy the goal of attribute unlearning. 
\noindent\textbf{Unlearning Speedup.}
Other than presenting superior performance on unlearning efficacy and fidelity retained after unlearning, our method also expect to achieve good performance with high efficiency and low memory footprint. Table~\ref{tab:efficiency} shows the speedup ratio and the unlearning latency per epoch before and after local parallelism acceleration. We can see that our method can speed up 1.42$\times$ meanwhile maintain prevailing performance after applying local parallelism. Further, we compare memory consumption for end-to-end and local propagation mode during the unlearning process. One can observe that local parallelism outperforms the end-to-end manner with 21.3\% reduction of memory requirements. As the datasets and models enlarge, more parallel blocks can be applied instead of two as the extension of our method for higher speed up ratio.
%To evaluate the training and inference efficiency, we conduct experiments on Fairface dataset with 180 training epoch. we also measure the parameters size and FLOPs with our methods on ResNet50. As shown in Tab.~\ref{tab:efficiency}, we can see that unlearning procedure do not introduce much overhead regarding the training and inference latency as well as model complexity, which can be a baseline level for further Vertical Machine Unlearning studies.

\noindent\textbf{Image Recovery Results.} We also visualize the effect of our method by reconstructing images given the feature map $\bm{h}$ after unlearning. We show the results of raw images, reconstruction before unlearning and reconstruction after unlearning from figure~\ref{fig:recon}. The ambiguity in certain area of the reconstructed images after unlearning is in sharp contrast to the one before unlearning, which is in accordance with the purpose and quantitative results of our methods.
%The ambiguity in certain area of the reconstructed images after unlearning makes it impossible to infer certain feature, i.e., age in this scenario, which is in accordance with the purpose and quantitative results of our methods.

%To further demonstrate superiority of model complexity. We evaluate the parameters size and FLOPs with our methods. As shown in Fig.~\ref{fig:complexity}, we achieves optimal parameters size and FLOPs by only reaches xx\% larger in parameter sizes and xx\% Flops in fulfilling the unlearning process.

% \begin{table*}[!h]
% \centering
% \begin{tabular}{l|l|l}
% \Xhline{3\arrayrulewidth}

% Models& Learning &Unlearning \\ 
%  \Xhline{3\arrayrulewidth}
% Training Latency /epoch(s)&  50.67  & \textbf{44.99} \\ \hline
    
% Inference Latency/epoch(s)& 0.423 &  \textbf{0.416}   \\\hline

% Parameters Size(M) & \textbf{23.512}& 23.574    \\ \hline

% Flops(G) & \textbf{0.6774}&  0.8375  \\ \Xhline{3\arrayrulewidth}

% \end{tabular}
% \vspace{3mm}
% \caption{\noindent\textbf{Unlearning efficiency and model complexity}: Training and inference latency of the unlearning process show more efficient compared to corresponding learning process, without introducing much overhead.  }
% \label{tab:efficiency}
% \end{table*}

\begin{table}[!h]
\centering
\begin{tabular}{l|l|l|l|l}
\Xhline{3\arrayrulewidth}

\multirow{2}*{Tasks}&Unlearnt&Efficacy &Fidelity & AUM\\ 
~&Attributes &($\downarrow$  \textit{better}) &($\uparrow$  \textit{better})&($\uparrow$  \textit{better})\\ \Xhline{3\arrayrulewidth}
\multirow{2}*{Gender}& Race & 0.1352  &    0.7950  &  0.8284   \\ 
~& Age &0.1729  &    0.8081   &  0.8175 \\ \hline
\multirow{3}*{Smiling}& Mustache & 0.5243 &   0.7874  &  0.5931  \\ 
~& Male & 0.5142 &  0.7830 & 0.5996\\
~ & Big Nose & 0.5267 & 0.7798 & 0.5891\\\hline
Living & Class &0.1095 & 0.9351 & 0.9123\\\Xhline{3\arrayrulewidth}

\end{tabular}
\vspace{1mm}
\caption{Ablation results on attribute type. Without loss of generality, we select different sensitive attributes for the purpose of considering the potential bias for certain attributes. We can see that they all reach similar high utility with sensitive attributes unlearned.}
\label{tab:ablation}
\end{table}

\subsection{Ablation Study}
\noindent\textbf{Attribute Sensitivity.}
To investigate the influence of attribute selection during unlearning process, we conduct ablation experiments to unlearn different attributes. Table~\ref{tab:ablation} shows the efficacy and fidelity result across different unlearnt attributes on ResNet18. One can see that despite unlearnt attributes are different, fidelity of main tasks remain stable and prevailing. Take the CelebA dataset as an example, we select mustache, male and big nose as the sensitive attributes from the smiling prediction model. We can tell that model performance remains comparative regardless of attribute type, or even competitive to the baseline in table~\ref{tab:utility}. We also consider the scenario of multi-attributes, where several attributes are unlearnt together in the process as the extension of our work.

\noindent\textbf{Hyper-parameter Sensitivity.} To study how the hyper-parameter $\lambda_1$, $\lambda_2$, $\lambda_3$ of section~\ref{subsec:Auxiliary} affect the performance, we change the value of hyper-parameters and calculate AUM for each combination. The results shown in figure~\ref{tab:ablation} imply that $\lambda_1$ plays an important role in the fidelity preservation. When $\lambda_1$ is too small, higher moderate $\lambda_2$ will contribute to achieving higher AMU. One can observe that optimal AMU can be obtained with relative larger $\lambda_1$ and adequate $\lambda_2$ and $\lambda 3$, which accords with our theorem and design .

%We can see that either attributes unlearning with prevailing performance with xx in age and xx in race. Further more, the utility has been largely preserved with xx and xx each. The ablation study in this dimension further prove the effectiveness of our method without the influence of selected features.

% \noindent\textbf{Unlearning with different position} 
% To further clear the variant of unlearning position in our method, we experiment on different positions of blocks within different models. For ResNet18, we range the position from block 3 to block 5, and all shows similar prevailing effects.  

% \begin{table}[!h]
% \centering
% %\resizebox{\columnwidth}{!}{%
% \begin{tabular}{l|l|l}
% \thickhline

% \multirow{2}*{Methods}&Training Time(s) &Inference Time(s) \\
% &/Speedup &/Speedup\\\thickhline

% Baseline& \textbf{22860.4}/-19.6\% &    11.3 /1.8\%   \\ \hline
% DisCo& 39125.4/37.6\%&   17.8/60.4\%   \\ \hline
% Sync-\SysName{}&33541.4/17.9\%  &  11.3 /1.8\%   \\\hline
% \SysName{}&28440.2/0\% & \textbf{11.1} /0\%      \\\thickhline

% \end{tabular}
% %}
% \caption{\noindent\textbf{Comparison of Execution Time on real Embedded Device}: Experiments results on Nvidia Xavier AGX demonstrate the reliability and robustness of \SysName{}.}
% \label{tab:edge_efficiency}
% \end{table}

\begin{figure}[t]
  \centering
  \includegraphics[width=1.0\linewidth]{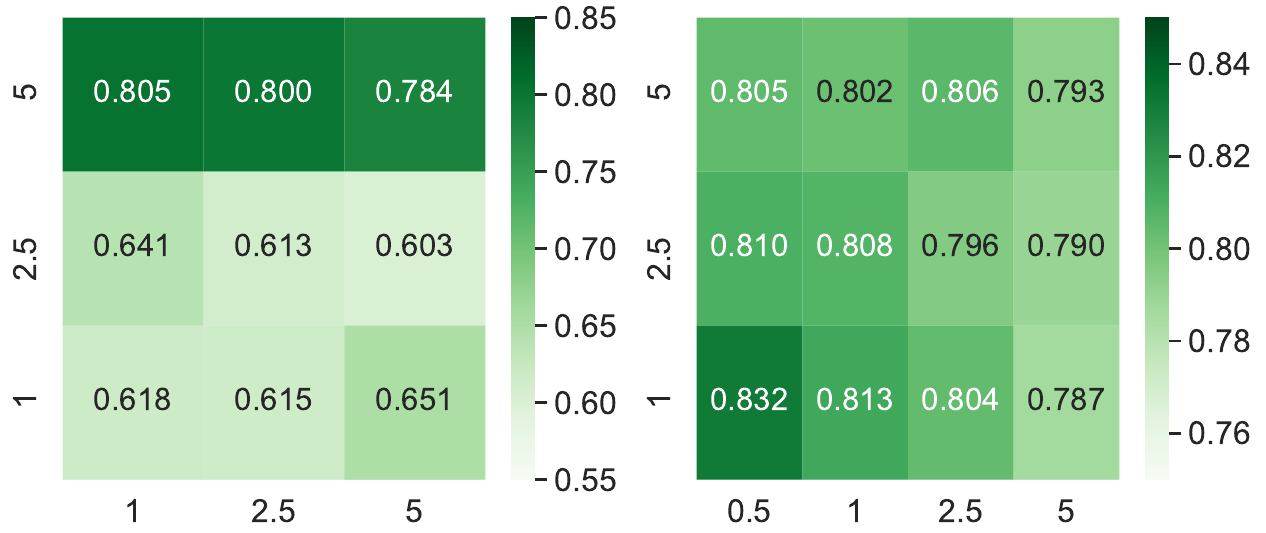}
  \caption{Hyper-parameter sensitivity analysis results. These hyper-parameters are as referred to section~\ref{subsec:Auxiliary}. We choose $\lambda_1$ as 5 (left) and 2.5 (right). The y-axis represents the scope of $\lambda_2$, and x-axis represents the scope of $\lambda_3$. We show the value of AUM across the combination of different hyper-parameters.}
  \label{fig:hyper}
  \vspace{2mm}
\end{figure}
\section{Related Work}\label{sec:Related}
The proposed method is related to two specific research works, the first one is current machine unlearning, and the second one is existing attribute-oriented task, i.e. fairness and de-biasing learning. Below we provide a brief review and discuss the main differences from those closest related researches.
\subsection{Machine Unlearning}
\noindent\textbf{Sample-Level Unlearning.}
Machine unlearning \cite{huang2021unlearnable,sekhari2021remember,gupta2021adaptive,brophy2021machine} was first brought up to solve the appeal of \textit{the right to be forgotten} raised in GDPR. The initial research focuses on the scenarios that data holders are entitled to revoke the data samples they no longer willing to share. Thus, it is conceivable that primary unlearning task is sample-level unlearning and concentrates on forgetting selected data samples and the contribution model memorized in the model. Although the concept has been early proposed be \cite{cao2015towards}, the first systematic work came up by \cite{bourtoule2021machine}, which strictly defines the problem of machine unlearning and introduces expedition strategy of retraining by ensemble learning, which establish a general method and a new baseline for further research. Later studies either focus on the different acceleration strategy with exact unlearning methods \cite{ginart2019making,brophy2021machine,schelter2021hedgecut}, or achieve higher efficiency with some appropriate unlearning methods by approximating the gradients or with historical data \cite{neel2021descent,izzo2021approximate,wu2020deltagrad}. 

\noindent\textbf{Other Unlearning.}
Other than focusing on efficiency problem of sample-level unlearning, recent studies set sights on the scenarios or object of machine unlearning \cite{chen2022recommendation,liu2022right,graves2021amnesiac,wang2022federated}. One of the recent arisen research tendency is class-level unlearning \cite{wang2022federated}, which targets on forgetting categories instead of random subset of data samples with channel pruning. Furthermore, client-level unlearning has been found needed in federated learning appropriately given the intrinsic architecture of FL and each client is entitled to revoke their data intentionally \cite{liu2022right}. Other unlearning tasks are tailor-made for each specific circumstance and requirement such as recommendation unlearning \cite{chen2022recommendation}, which aims at erasing the impacts of their sensitive data from the trained recommendation system.    

\subsection{Fairness and De-biasing Learning }
Unlike traditional unlearning task, we orientate our works as a attribute-level unlearning, so we need to compare our work with existing researches related to attribute problems. With the development of artificial intelligent, neural network models have largely engaged in human daily lives. AI systems are employed in critical decision making tasks, e.g., criminal sentencing, face recognition, and risk assessments, where fairness is of great importance. However, bias and discrimination in decision making system lie in several ways \cite{singh2022anatomizing}, especially in the historical training data. For example, \cite{nam2020learning} states that attributes that `easier' to learn than the desired knowledge is likely to become bias, e.g., texture, color. Instead of learning the decision rule based on the object in the images, neural networks often learn to make predictions with unintended rule when dealing with datasets with bias. Another problem happens in criminal sentencing or risk assessments systems where race and gender are usually become bias and discrimination causing by the historical training trace. The key aim is to make fair decisions and obtain both group and individual fairness. 

Some researches deal with these issue by oversampling techniques \cite{bickel2009discriminative,elkan2001foundations}, duplicating minority samples in imbalanced data and give them higher weight during the training process. Some works attempt to learn separate classifiers for each bias attribute \cite{ryu2017inclusivefacenet,wang2020towards}. Other researches improve fairness through a form of fairness regularization to achieve statistical parity \cite{nam2020learning}. Recently, with the widespread application of Generative
Adversarial Network \cite{goodfellow2014generative}, several works use GAN to augment biased real-world datasets with multiple target labels and protected attributes \cite{ramaswamy2021fair}.
\section{Conclusion}\label{sec:Conclusion}
%should not mention vertical learning here 

In this paper, we have proposed a novel machine unlearning mechanism that can selectively remove sensitive attributes from latent feature space. 
We deem the proposed attribute unlearning paradigm a new initiative for responding to the emergent privacy protection legislation. We have defined the concept of attribute unlearning and elaborated the objectives to protect sensitive attribute information from trained models. A representation detachment method has been developed to progressively separate sensitive attributes from feature representations that are kept for the classification tasks by dedicating a representation detachment training. To expedite such loss calculation, we have introduced an approximation method for the loss value with rigorous theoretical analysis. Extensive experiments have been conducted to verify our theory and performance. The proposed method has shown a new direction of protecting personal information and made an important step forward to the research and development of machine unlearning mechanisms.

% In this paper, we have proposed a novel insight of machine unlearning mechanism by selectively removing sensitive attributes from latent feature space, which can be regarded as a novel interpretation for emergent privacy protection legislation, namely, the vertical machine unlearning, a new research initiative for machine unlearning. We have defined the concept and elaborated the purpose of vertical machine unlearning to protect sensitive feature information from trained models. A representation detachment method have been proposed to progressively separate sensitive features from intermediate features which are used for final classification, via a deliberately designing the representation detachment loss. Moreover, to address the difficulties in calculating above loss, we have introduced an approximation method for the loss value with rigorous theoretical analysis. Our method have shown a new direction of protecting personal information and made an important step forward to the research and development of machine unlearning mechanisms. 

\newpage

\bibliographystyle{ACM-Reference-Format}
\bibliography{reference}

% \appendix

\end{document}

\title{Supplementary material}
\maketitle

\section{PROOF OF PROPOSITION 1}

\begin{proposition} \label{approximate loss}
Suppose that the Markov chain $(y,r, s)\rightarrow \boldsymbol{x}\rightarrow \boldsymbol{h}$ holds. Then the upper bound of $\mathcal{L}_{infoFiltra}$ is as follows
\begin{eqnarray}
    \mathcal{L}_{infoFiltra} &\le& -\lambda_1I(\boldsymbol{h},\boldsymbol{x})-\lambda_2I(\boldsymbol{h},y)-\lambda_3I(\boldsymbol{h},z) \nonumber \\
    &=&\overline{\mathcal{L}}_{infoFiltra}
\end{eqnarray}
where $\lambda_1=\alpha(1-\beta)$, $\lambda_2=\alpha\beta$, $\lambda_3=\alpha(\beta-\gamma)$, $\beta\geq\gamma$, and $z$ represent the sensitive attribute.
\end{proposition}

\begin{proof}
Note that, our original loss function is given by
\begin{align}
    \mathcal{L}_{infoFiltra} =& \alpha(-I(\boldsymbol{h},\boldsymbol{x})+\beta I(\boldsymbol{h}, r^*)+\gamma I(\boldsymbol{h}, s^*)),\\
    \mathrm{s.t.}r^*=& argmax_{r,I(r,\boldsymbol{x})>0, I(r,y)=0}I(\boldsymbol{h},r),\notag\\
    s^*=& argmax_{s, I(s, \boldsymbol{x})>0, I(s,y)=0}I(\boldsymbol{h},s). \notag
\end{align} 
where $\alpha, \beta, \gamma \geq 0$. Since the Markov chain $(y,r, s)\rightarrow \boldsymbol{x}\rightarrow \boldsymbol{h}$, we have the following inequality
\begin{equation}\label{eq:markov}
    I(\boldsymbol{h}, (y,r^*,s^*)) \le I(\boldsymbol{h}, \boldsymbol{x})
\end{equation}
Given that
\begin{align} \label{eq:prop}
    &\quad I(\boldsymbol{h}, (y,r^*,s^*))\notag\\
    &=H(\boldsymbol{h})-H(\boldsymbol{h}|y, r^*, s^*)\notag \\
    &=H(\boldsymbol{h})-H(\boldsymbol{h}|r^*)+H(\boldsymbol{h}|r^*)-H(\boldsymbol{h}|y, r^*, s^*)\notag\\
    &=I(\boldsymbol{h}, r^*)+I(\boldsymbol{h},(y,s^*)|r^*)
\end{align}
Where $H(\cdot)$ represents information entropy and mutual information has the property that $I(a,b)=H(a)-H(a|b)$. Considering that $y, r^*$ and $s^*$ are independent of each other, and the property of mutual information where $I(a,b)=I(b,a)$, then
\begin{align}\label{eq:independent}
    &\quad I(\boldsymbol{h},(y,s^*)|r^*) \notag\\
    &= H((y,s^*)|r^*)-H((y,s^*)|\boldsymbol{h}, r^*) \notag \\
    &\geq H(y, s^*)-H(y, s^*|\boldsymbol{h}) ]\notag \\
    &=I((y,s^*), \boldsymbol{h}) \notag \\
    &=H(\boldsymbol{h})-H(\boldsymbol{h}|y,s^*) \notag \\
    &=H(\boldsymbol{h})-H(\boldsymbol{h}|s^*)+H(\boldsymbol{h}|s^*)-H(\boldsymbol{h}|y,s^*) \notag \\
    &=I(\boldsymbol{h},s^*)+I(\boldsymbol{h},y|s^*) \notag \\
    &=I(\boldsymbol{h},s^*)+H(y|s^*)-H(y|\boldsymbol{h},s^*) \notag\\
    &\geq I(\boldsymbol{h},s^*)+H(y)-H(y|\boldsymbol{h}) \notag \\
    &=I(\boldsymbol{h},s^*)+I(\boldsymbol{h},y)
\end{align}
Combining the equation \ref{eq:prop} and the inequality \ref{eq:markov}, \ref{eq:independent}, we can get the following inequality
\begin{equation}
    I(\boldsymbol{h},s^*)+I(\boldsymbol{h},r^*)\le I(\boldsymbol{h},\boldsymbol{x})-I(\boldsymbol{h},y)
\end{equation}
So we can eliminate $I(\boldsymbol{h}, r^*)$ in the original loss function that are hard to calculate.
\begin{align}
    &\quad\mathcal{L}_{infoFiltra}\notag\\
    &=\alpha(-I(\boldsymbol{h},\boldsymbol{x})+\beta(I(\boldsymbol{h},r^*)+I(\boldsymbol{h}, s^*))+p) \notag\\
    &\le\alpha(-I(\boldsymbol{h},\boldsymbol{x})+\beta(I(\boldsymbol{h},\boldsymbol{x})-I(\boldsymbol{h},y))+p)
\end{align}
Where $p=(\gamma-\beta)I(\boldsymbol{h},s^*)$. Then our target is to find the bound for $p$. We can deduce sensitive information from sensitive attributes. There is another Markov chain $z\rightarrow s$. Then $I(\boldsymbol{h},z)\le I(\boldsymbol{h},s^*)$. Combining the assumption $\beta \geq \gamma$, the upper bound of $p$ is $p\le(\gamma-\beta)I(\boldsymbol{h},z)$. We have already discussed the necessity of $\beta \geq \gamma$ in the text. So we can prove that
\begin{align}
    \mathcal{L}_{infoFiltra}&\le\alpha(-I(\boldsymbol{h},\boldsymbol{x})+\beta(I(\boldsymbol{h},\boldsymbol{x})-I(\boldsymbol{h},y))\notag \\
    &\quad+(\gamma-\beta)I(\boldsymbol{h},z)) \notag \\
    &=-\lambda_1I(\boldsymbol{h},\boldsymbol{x})-\lambda_2I(\boldsymbol{h},y)-\lambda_3I(\boldsymbol{h},z)
\end{align}
where $\lambda_1=\alpha(1-\beta)$, $\lambda_2=\alpha\beta$, $\lambda_3=\alpha(\beta-\gamma)$.
\end{proof}

\section{PROOF OF PROPOSITION 2}
\begin{proposition}\label{gap}
Assume that there exists a deterministic function map $x$ to $y$, the gap $\epsilon=\overline{\mathcal{L}}_{infoFiltra}-\mathcal{L}_{infoFiltra}$ can be bounded by
\begin{equation}
    \epsilon \le \alpha\beta(I(\boldsymbol{x},y)-I(\boldsymbol{h},y)-I(\boldsymbol{h},z))
\end{equation}
\end{proposition}
\begin{proof}
The gap $\epsilon$ is given by
\begin{eqnarray}
    \epsilon &=& \alpha(\beta(\underbrace {I(\boldsymbol{h},\boldsymbol{x})-I(\boldsymbol{h},y)-I(\boldsymbol{h},r^*)}_{Q_1})+ \notag \\
    &\quad&\gamma(\underbrace{I(\boldsymbol{h},z)-I(\boldsymbol{h},s^*)}_{Q_2})-\beta I(\boldsymbol{h},z))
\end{eqnarray}
Suppose there is a function $f$ make that $\boldsymbol{x}=f(y, r, s)$. Let $\widetilde{r}$ and $\widetilde{s}$ be the random variable in the function $f$. We have
\begin{eqnarray}
    I(\boldsymbol{h}, \boldsymbol{x})&=&I(\boldsymbol{h},(y, \widetilde{r}, \widetilde{s}))\notag\\
    &=&I(\boldsymbol{h},\widetilde{r})+I(\boldsymbol{h},(y,\widetilde{s})|\widetilde{r})
\end{eqnarray}
Obviously, $I(\boldsymbol{h}, \widetilde{r})\le I(\boldsymbol{h}, r^*)$. We obtain
\begin{eqnarray}
    Q_1 &\le& I(\boldsymbol{h},\boldsymbol{x})-I(\boldsymbol{h},y)-I(\boldsymbol{h},\widetilde{r}) \notag \\
    &=& I(\boldsymbol{h},\widetilde{r})+I(\boldsymbol{h},(y,\widetilde{s})|\widetilde{r})-I(\boldsymbol{h},y)-I(\boldsymbol{h},\widetilde{r}) \notag \\
    &=&I(\boldsymbol{h},(y,\widetilde{s})|\widetilde{r})-I(\boldsymbol{h},y) \notag \\
    &=&I(\boldsymbol{h},(y,\widetilde{s})|\widetilde{r})-I(\boldsymbol{x},y)+I(\boldsymbol{x},y)-I(\boldsymbol{h},y) \notag \\
    &=&H((y,\widetilde{s})|\widetilde{r})-H((y,\widetilde{s})|\boldsymbol{h},\widetilde{r})-H(y)+H(y|\boldsymbol{x})\notag \\
    &\quad&+I(\boldsymbol{x},y)-I(\boldsymbol{h},y)
\end{eqnarray}
Given that $y, \widetilde{s}$ and $\widetilde{r}$ are independent of each other, we have $H((y,\widetilde{s})|\widetilde{r})=H(y,\widetilde{s})$, and $y$ can be derived from $x$ so that $H(y|\boldsymbol{x})=0$. We get
\begin{equation}
    Q_1\le I(\boldsymbol{x},y)-I(\boldsymbol{h},y)
\end{equation}
Due to $Q_2\le0$, we can prove that
\begin{equation}
    \epsilon \le \alpha\beta(I(\boldsymbol{x},y)-I(\boldsymbol{h},y)-I(\boldsymbol{h},z))
\end{equation}
\end{proof}

\title{Appendix}
\maketitle

\section{Proof of Lemma~\ref{approximate loss}}

\begin{lemma} \label{approximate loss}
Suppose that the Markov chain $(y,s)\rightarrow \boldsymbol{x}\rightarrow \boldsymbol{h}$ holds. Then the upper bound of $\mathcal{L}$ is as follows
\begin{eqnarray}
    \mathcal{L} \le -\lambda_1I(\boldsymbol{h},\boldsymbol{x})-\lambda_2I(\boldsymbol{h},y)-\lambda_3I(\boldsymbol{h},z) = \overline{\mathcal{L}},
\end{eqnarray}
where $\lambda_1=\alpha(1-\beta)$, $\lambda_2=\alpha\beta$, $\lambda_3=\alpha(\beta-\gamma)$, $\beta\geq\gamma$, and $z$ represents the sensitive attribute.
\end{lemma}

\begin{proof}
Note that, our original loss function is given by
\begin{align}
    \mathcal{L} =& \alpha\big[-I(\boldsymbol{h},\boldsymbol{x})+\gamma I(\boldsymbol{h}, \hat{s})\big], \notag \\
    \mathrm{s.t.} \quad 
    \hat{s}=& {\arg_{s}\max}_{I(s, \boldsymbol{x})>0, I(s,y)=0}I(\boldsymbol{h},s), 
\end{align} 
where $\alpha, \gamma \geq 0$. Since the Markov chain $(y,s)\rightarrow \boldsymbol{x}\rightarrow \boldsymbol{h}$, we have the following inequality
\begin{equation}\label{eq:markov}
    I\big(\boldsymbol{h}, (y,\hat{s})\big) \le I(\boldsymbol{h}, \boldsymbol{x}).
\end{equation}
Given that
\begin{align} \label{eq:prop}
    \quad I\big(\boldsymbol{h}, (y,\hat{s})\big)
    &=H(\boldsymbol{h})-H(\boldsymbol{h}|y, \hat{s})\notag \\
    &=H(\boldsymbol{h})-H(\boldsymbol{h}|\hat{s})+H(\boldsymbol{h}|\hat{s})-H(\boldsymbol{h}|y, \hat{s}) \\
    &=I(\boldsymbol{h}, \hat{s})+I(\boldsymbol{h},y|\hat{s}), \notag
\end{align}
where $H(\cdot)$ represents information entropy and mutual information has the property that $I(a,b)=H(a)-H(a|b)$. Considering that $y$ and $\hat{s}$ are independent of each other, and the property of mutual information where $I(a,b)=I(b,a)$, then
\begin{align}\label{eq:independent}
    \quad I(\boldsymbol{h},y|\hat{s}) 
    &= H(y|\hat{s})-H(y|\boldsymbol{h}, \hat{s}) \notag \\
    &\geq H(y, \hat{s})-H(y, \hat{s}|\boldsymbol{h}) \notag \\
    &=I\big((y,\hat{s}), \boldsymbol{h}\big) \notag \\
    &=H(\boldsymbol{h})-H(\boldsymbol{h}|y,\hat{s}) \notag \\
    &=H(\boldsymbol{h})-H(\boldsymbol{h}|\hat{s})+H(\boldsymbol{h}|\hat{s})-H(\boldsymbol{h}|y,\hat{s}) \\
    &=I(\boldsymbol{h},\hat{s})+I(\boldsymbol{h},y|\hat{s}) \notag \\
    &=I(\boldsymbol{h},\hat{s})+H(y|\hat{s})-H(y|\boldsymbol{h},\hat{s}) \notag\\
    &\geq I(\boldsymbol{h},\hat{s})+H(y)-H(y|\boldsymbol{h}) \notag \\
    &=I(\boldsymbol{h},\hat{s})+I(\boldsymbol{h},y). \notag
\end{align}
Combining the equation \ref{eq:prop} and the inequality \ref{eq:markov}, \ref{eq:independent}, we can get the following inequality
\begin{equation}
    I(\boldsymbol{h},\hat{s})\le I(\boldsymbol{h},\boldsymbol{x})-I(\boldsymbol{h},y).
\end{equation}
So we can eliminate $I(\boldsymbol{h}, \hat{s})$ in the original loss function that are hard to calculate.
\begin{align}
    \quad\mathcal{L}&=\alpha\big[-I(\boldsymbol{h},\boldsymbol{x})+\beta I(\boldsymbol{h}, \hat{s})+p\big] \notag\\
    &\le\alpha\Big[-I(\boldsymbol{h},\boldsymbol{x})+\beta\big(I(\boldsymbol{h},\boldsymbol{x})-I(\boldsymbol{h},y)\big)+p\Big],
\end{align}
where $p=(\gamma-\beta)I(\boldsymbol{h},\hat{s})$. Then our target is to find the bound for $p$. We can deduce sensitive information from sensitive attributes. There is another Markov chain $z\rightarrow s$. Then $I(\boldsymbol{h},z)\le I(\boldsymbol{h},\hat{s})$. Combining the assumption $\beta \geq \gamma$, the upper bound of $p$ is $p\le(\gamma-\beta)I(\boldsymbol{h},z)$. So we can prove that
\begin{align}
    \mathcal{L}&\le\alpha\Big[-I(\boldsymbol{h},\boldsymbol{x})+\beta\big(I(\boldsymbol{h},\boldsymbol{x})-I(\boldsymbol{h},y)\big)+(\gamma-\beta)I\big(\boldsymbol{h},z)\Big] \notag \\
    &=-\lambda_1I(\boldsymbol{h},\boldsymbol{x})-\lambda_2I(\boldsymbol{h},y)-\lambda_3I(\boldsymbol{h},z)
\end{align}
where $\lambda_1=\alpha(1-\beta)$, $\lambda_2=\alpha\beta$, $\lambda_3=\alpha(\beta-\gamma)$.
\end{proof}

\section{Proof of Lemma~\ref{gap}}
\begin{lemma} \label{gap}
Assume there exists a deterministic function map $x$ to $y$, the gap $\epsilon=\overline{\mathcal{L}}-\mathcal{L}$ is bounded by
\begin{equation}
    \epsilon \le \alpha\beta\big[I(\boldsymbol{x},y)-I(\boldsymbol{h},y)-I(\boldsymbol{h},z)\big]
\end{equation}
\end{lemma}
\begin{proof}
The gap $\epsilon$ is given by
\begin{eqnarray}
    \epsilon= \alpha\Big[\beta\big(\underbrace {I(\boldsymbol{h},\boldsymbol{x})-I(\boldsymbol{h},y)}_{Q_1}\big)+ \gamma\big(\underbrace{I(\boldsymbol{h},z)-I(\boldsymbol{h},\hat{s})}_{Q_2}\big)-\beta I(\boldsymbol{h},z)\Big]
\end{eqnarray}
Suppose there is a function $f$ make that $\boldsymbol{x}=f(y,s)$. Let $\widetilde{s}$ be the random variable in the function $f$. We have
\begin{align}\label{function}
    I(\boldsymbol{h}, \boldsymbol{x})&\le I\big(\boldsymbol{h},(y, \widetilde{s})\big)
\end{align}
Obviously, according to equation~\ref{function}. We obtain
\begin{align}
    Q_1 &= I(\boldsymbol{h},\boldsymbol{x})-I(\boldsymbol{h},y) \notag \\
    &\le I\big(\boldsymbol{h},(y,\widetilde{s})\big)-I(\boldsymbol{h},y) \notag \\
    &=I\big(\boldsymbol{h},(y,\widetilde{s})\big)-I(\boldsymbol{x},y)+I(\boldsymbol{x},y)-I(\boldsymbol{h},y) \\
    &=H(y,\widetilde{s})-H\big((y,\widetilde{s})|\boldsymbol{h}\big)-H(y)+H(y|\boldsymbol{x})+I(\boldsymbol{x},y)-I(\boldsymbol{h},y) \notag
\end{align}
Given that $y, \widetilde{s}$ are independent of each other, and $y$ can be derived from $x$ so that $H(y|\boldsymbol{x})=0$. We get
\begin{equation}
    Q_1\le I(\boldsymbol{x},y)-I(\boldsymbol{h},y)
\end{equation}
Due to $Q_2\le0$, we can prove that
\begin{equation}
    \epsilon \le \alpha\beta\big[I(\boldsymbol{x},y)-I(\boldsymbol{h},y)-I(\boldsymbol{h},z)\big]
\end{equation}
\end{proof}